%% file: main.tex
\documentclass[10pt,twocolumn,letterpaper]{article}

 \usepackage[pagenumbers]{cvpr} 

\input{preamble}
\definecolor{cvprblue}{rgb}{0.21,0.49,0.74}
\usepackage{amssymb}
\usepackage{bm}
\usepackage{afterpage}
\usepackage{float}
\usepackage{amsmath}
\usepackage{graphicx}
\usepackage{multirow}

\newcommand{\myparagraph}[1]{
	\vspace{0.1cm}\noindent
	\textbf{#1}
}

\usepackage[pagebackref,breaklinks,colorlinks,allcolors=cvprblue]{hyperref}

\def\paperID{1723} 
\def\confName{CVPR}
\def\confYear{2026}

\title{ResDiT: Evoking the Intrinsic Resolution Scalability in Diffusion Transformers}

\author{
	Yiyang Ma$^{1*}$ \quad Feng Zhou$^{1*}$ \quad Xuedan Yin$^{2}$ \quad Pu Cao$^{1}$ \quad Yonghao Dang$^{1}$ \quad Jianqin Yin$^{1\dagger}$ \\
	$^{1}$Beijing University of Posts and Telecommunications \quad $^{2}$Tsinghua University \\
	{\tt\small \{yym2024,zhoufeng,caopu,dyh2018,jqyin\}@bupt.edu.cn, yxd23@mails.tsinghua.edu.cn} \\
	\small$^*$Equal contribution. \quad $^\dagger$Corresponding author.
}

\begin{document}
	
\twocolumn[{
	\renewcommand\twocolumn[1][]{#1}
	\begin{center}
		\centering
		\maketitle
		\includegraphics[width=0.9\textwidth, keepaspectratio]{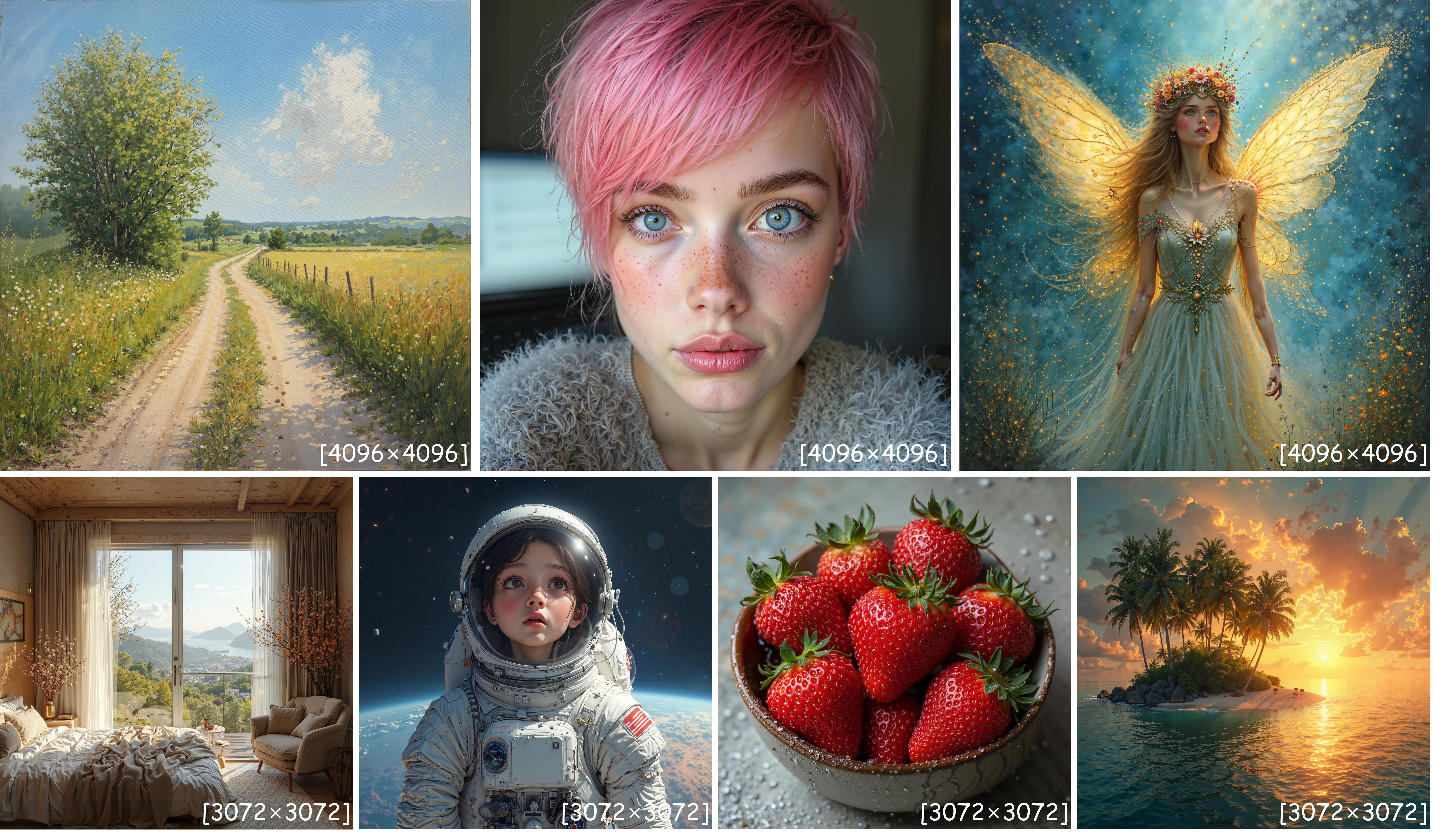}
		\captionof{figure}{\textbf{Qualitative examples of the proposed ResDiT,} which enables the pre-trained T2I models to generate high-resolution images than the originally trained resolution, without any training or fine-tuning. \textbf{Best view ZOOM-IN.}}
		\label{fig:fig1}
	\end{center}
}]

\input{sec/0_abstract}    
\input{sec/1_intro}

\input{sec/2_relatedwork}
\input{sec/3_method}

\input{sec/4_exp}

{
    \small
    \bibliographystyle{ieeenat_fullname}
    \bibliography{main}
}
\clearpage
\input{sec/cvpr_sup}

\end{document}

%% file: sec/0_abstract.tex
\begin{abstract}

Leveraging pre-trained Diffusion Transformers (DiTs) for high-resolution (HR) image synthesis often leads to spatial layout collapse and degraded texture fidelity. Prior work mitigates these issues with complex pipelines that first perform a base-resolution (i.e., training-resolution) denoising process to guide HR generation. We instead explore the intrinsic generative mechanisms of DiTs and propose ResDiT, a training-free method that scales resolution efficiently. We identify the core factor governing spatial layout, position embeddings (PEs), and show that the original PEs encode incorrect positional information when extrapolated to HR, which triggers layout collapse. To address this, we introduce a PE scaling technique that rectifies positional encoding under resolution changes. To further remedy low-fidelity details, we develop a local-enhancement mechanism grounded in base-resolution local attention. We design a patch-level fusion module that aggregates global and local cues, together with a Gaussian-weighted splicing strategy that eliminates grid artifacts. Comprehensive evaluations demonstrate that ResDiT consistently delivers high-fidelity, high-resolution image synthesis and integrates seamlessly with downstream tasks, including spatially controlled generation.

%
%
\end{abstract}

%% file: sec/1_intro.tex
\begin{figure*}[htbp]
	\centering
	\includegraphics[width=0.9\textwidth,height=0.5\textheight, keepaspectratio]{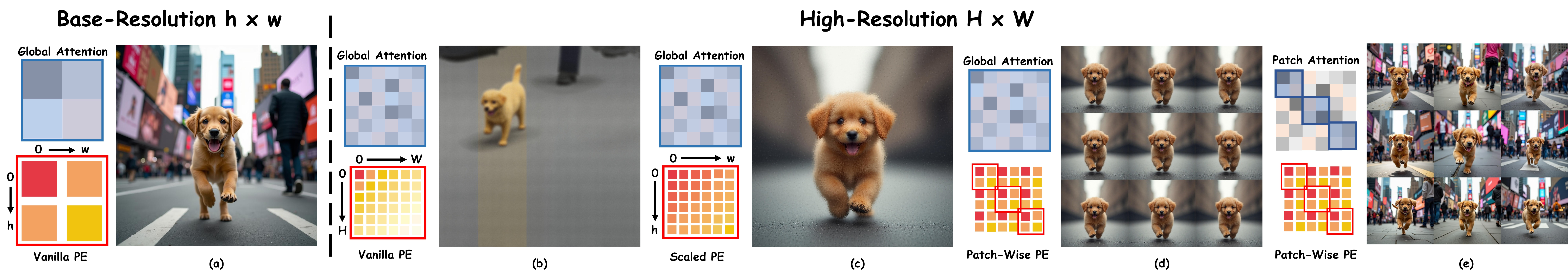}
	\caption{\textbf{Disentangling PE and attention range in high-resolution DiT synthesis.} Systematic interventions on positional embeddings (PEs) and attention range across resolutions. (a) At base resolution, a DiT with global attention and vanilla PE produces coherent layouts and fine details. (b) When directly applied to high resolution, layout collapse occurs as the subject becomes shrunken and misplaced due to a mismatch between PE and the attention field. (c) Using a scaled PE restores spatial arrangement but yields blurred details. (d) Applying patch-wise base-resolution PEs ensures correct local structure within each patch, yet details remain degraded. (e) Introducing patch-level local attention further enhances fine details. These results show that positional embeddings determine spatial arrangement, while the attention receptive-field scale governs detail fidelity in DiTs.}
	\label{fig:fig2}
\end{figure*}

\section{Introduction}
\label{sec:intro}

Text-to-image generation models have garnered widespread attention due to their impressive capabilities and wide range of applications, such as customized content creation and individualized media synthesis~\cite{cao2024controllablegenerationtexttoimagediffusion,poole2022dreamfusion}. Recently, a new generation of state-of-the-art models, such as the Flux~\cite{flux2024} and SD3~\cite{esser2024scaling},  has established a powerful paradigm for high-fidelity image synthesis by introducing transformer architectures~\cite{peebles2023scalable} that capture long-range dependencies and enable scalable global modeling to support high-quality image generation. However, these models often struggle to generate images beyond their training resolution, with high-resolution inference leading to noticeable degradation or even complete failure~\cite{bu2025hiflow}. This limitation constrains their applicability in tasks that require high-resolution outputs.

A straightforward solution is to train or fine-tune models directly at high resolutions~\cite{guo2024make,hoogeboom2023simple,liu2024linfusion,ren2024ultrapixel,xie2023difffit}. This approach demands high-quality, high-resolution datasets and incurs significant computational costs. As a result, various training-free methods have been proposed as more efficient alternatives~\cite{jin2023training,he2023scalecrafter,huang2024fouriscale,du2024demofusion,bar2023multidiffusion,zhou2025exploring,lin2024accdiffusion,kim2025diffusehigh,cao2024ap,yang2025fam,zhang2024frecas,wu2025megafusion}. Some methods are tailored to specific model architectures (U-Net) and do not readily generalize to DiT-based models~\cite{he2023scalecrafter,huang2024fouriscale,zhou2025exploring}. Other methods adopt a two-stage paradigm, where a base-resolution image is first generated and then used to guide high-resolution synthesis~\cite{du2024demofusion,kim2025diffusehigh,du2024max,bu2025hiflow}. While effective, it heavily depends on the base-resolution denoising image and introduces unnecessary complexity. More importantly, they fundamentally treat high-resolution generation as a super-resolution task, relying on external guidance rather than unlocking the model’s intrinsic capability to generate high-resolution content.

In this work, we investigate how to address two major challenges in high-resolution synthesis, layout collapse and degraded details, from an intrinsic, mechanics-based perspective. Since attention serves as the key spatial mechanism enabling token interactions in DiTs, we begin by analyzing two crucial spatial factors within attention: positional embeddings (PEs)~\cite{su2024roformer} and the attention range field. As illustrated in \cref{fig:fig2}, we systematically intervene on these components under different resolutions.
In the base-resolution setting (\cref{fig:fig2}(a)), a DiT equipped with global attention and vanilla PE at resolution $h \times w$ generates images with both coherent global layouts and high-quality details. When we directly apply the same global attention and vanilla PE to a high-resolution latent $H \times W$ at test time (\cref{fig:fig2}(b)), the model suffers from layout collapse: the main subject becomes shrunken and misplaced, revealing a clear mismatch between the extrapolated PE and the expanded attention field.
To decouple these factors, we keep the global attention range at high resolution but replace the HR latent’s PE with its base-resolution counterpart (\cref{fig:fig2}(c)). This “scaled PE’’~\cite{chen2023extending} operation largely restores correct spatial arrangement, yet the generated details remain blurry and low-fidelity.
We then introduce a patch-wise base-resolution PE that is tiled across the HR canvas (\cref{fig:fig2}(d)). Experiments show that PE primarily governs the spatial arrangement of objects, but a mismatch between the expanded attention field and the one used during training still leads to detail loss. This observation is further confirmed in \cref{fig:fig2}(e): once we adopt patch-level local attention~\cite{liu2021swin} (effectively generating multiple regions independently), the image details improve substantially. These findings lead to an important mechanistic insight about DiTs: positional embeddings dictate spatial arrangement, while the attention receptive-field scale critically determines the model’s ability to generate high-quality details.

Building on these observations, we introduce ResDiT, a training-free framework for high-resolution image synthesis. To improve spatial arrangement at high resolutions, we employ a scaled positional embedding~\cite{chen2023extending} mechanism that maps HR positional embeddings back to the training resolution, ensuring that spatial relationships are generated correctly. Meanwhile, to enhance local detail quality, we propose an overlapping patch partitioning and splicing strategy. This design enables local attention~\cite{liu2021swin} computation alongside the original global attention, and uses Gaussian-weighted splicing in overlapping regions to suppress grid artifacts during synthesis. To further integrate the strengths of the two attention receptive fields, we introduce a patch-wise spectral fusion technique that preserves high-frequency components from patch attention—which carry fine details—while retaining the low-frequency components from global attention, which encode coherent spatial structure.

In summary, our contributions are as follows:

\begin{itemize}
	\item We provide a mechanistic analysis of DiTs under high-resolution inference. Through controlled interventions on positional embeddings and attention range, we reveal that PE determines spatial arrangement, while attention receptive-field mismatch causes detail degradation. This explains the root causes of layout collapse and low-fidelity details in HR synthesis.
	\item We propose ResDiT, a fully training-free method for high-resolution synthesis in DiTs. ResDiT corrects spatial arrangement via scaled positional embeddings and restores details through overlapping patch attention with Gaussian splicing. A patch-wise spectral fusion module further combines global structure with local detail fidelity.
	\item We perform extensive experiments and ablation studies on high-resolution image synthesis, showcasing the capability of our approach. We also integrate ResDiT with an off-the-shelf spatial-control method, showing the compatibility and practical utility.
\end{itemize}

%% file: sec/2_relatedwork.tex
\section{Related Work}
\label{sec:relatedwork}

\subsection{Text-to-Image Synthesis}
Diffusion models~\cite{ho2020denoising} have become the dominant paradigm for text-to-image (T2I) synthesis, largely replacing GAN-based~\cite{goodfellow2020generative} approaches due to their stability and ability to generate high-fidelity images from complex prompts. Early T2I diffusion systems primarily relied on U-Net backbones~\cite{ronneberger2015u}, and latent diffusion~\cite{rombach2022high} models such as Stable Diffusion~\cite{podell2023sdxl} further improved efficiency by performing denoising in a learned latent space, enabling high-resolution generation at manageable computational cost. More recent models move beyond the U-Net architecture and adopt Diffusion Transformers (DiTs)~\cite{peebles2023scalable}, which treat images as sequences of latent tokens and scale more effectively. SD3~\cite{esser2024scaling} exemplifies this shift by fully transitioning to a DiT-based design, while FLUX~\cite{flux2024} and its variants extend this direction with large multimodal Transformer~\cite{esser2024scaling} blocks and flow-based~\cite{liu2022flow,lipman2022flow} training objectives. These advances yield stronger prompt fidelity, richer fine-grained detail, and more flexible resolution control. Our work targets this family of Transformer-based T2I models and enables training-free resolution scaling for SD3 and FLUX.

\subsection{Training-free High-Resolution Image Synthesis}
Despite considerable advances in image synthesis, generating high-resolution images (e.g., 3K and above) remains challenging. Training-free methods~\cite{jin2023training,he2023scalecrafter,huang2024fouriscale,du2024demofusion,bar2023multidiffusion,zhou2025exploring,lin2024accdiffusion,kim2025diffusehigh,cao2024ap,yang2025fam,zhang2024frecas,wu2025megafusion} have emerged as a promising direction, as they leverage powerful pre-trained diffusion models and avoid the substantial data and computational costs of re-training at higher resolutions. In the U-Net–based diffusion regime, several training-free approaches address high-resolution generation within a single denoising process: ScaleCrafter~\cite{he2023scalecrafter} enlarges the receptive field via dilated convolutions, while PBC~\cite{zhou2025exploring} introduces virtual zero-padded boundaries to provide correct spatial context, yielding more coherent global structures. However, these techniques are closely tied to convolutional U-Net architectures and do not directly transfer to DiT-based models.
With the shift toward Diffusion Transformers, training-free high-resolution methods have largely adopted two-stage pipelines: they first perform a base-resolution generation and then use this trajectory to guide sampling at higher resolutions. For example, I-Max~\cite{du2024max} models base-resolution flows as projections of high-resolution flows and exploits the linear interpolation property of rectified flows to construct dynamic guidance that steers the high-resolution trajectory. HiFlow~\cite{bu2025hiflow}, similarly, derives a virtual reference flow from base-resolution information to regularize high-resolution sampling and fuse coarse structure with fine details. While effective, these methods inherently rely on base-resolution guidance and introduce additional complexity into the denoising process.

%% file: sec/3_method.tex
\begin{figure*}[htbp]
	\centering
	\includegraphics[width=0.85\textwidth, keepaspectratio]{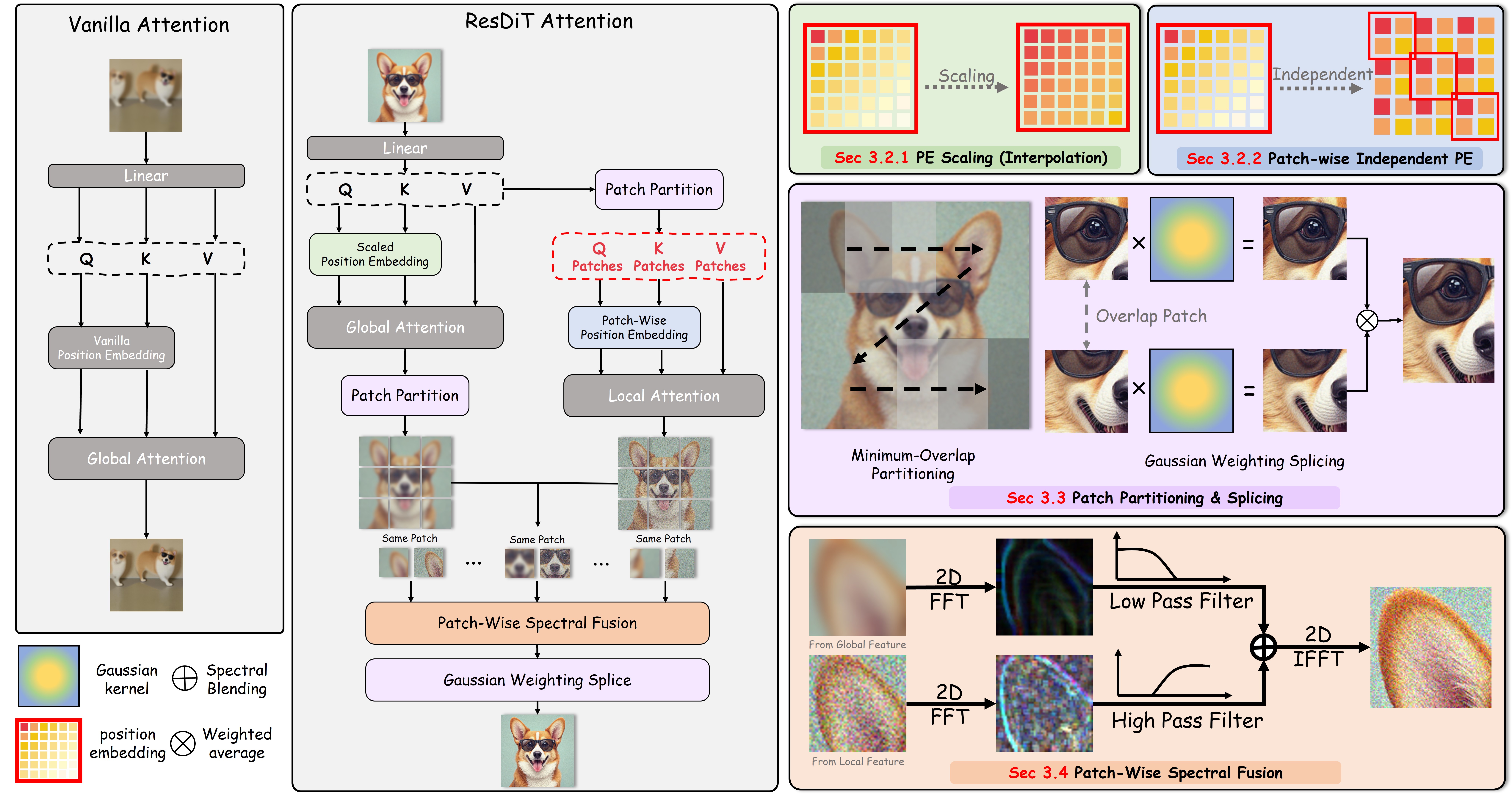}
	\caption{\textbf{Overview of ResDiT.} ResDiT restructures the vanilla attention mechanism in Diffusion Transformers (DiTs) into two complementary branches to enable training-free resolution scaling. Specifically, the global branch performs global attention with scaled positional embeddings to preserve the overall spatial layout, while the local branch applies patch-level attention to enhance fine-grained details. To maintain continuity across patches, we propose a Minimum-Overlap Partitioning strategy that ensures contextual consistency at patch boundaries and a Gaussian Weighting Splicing scheme that smoothly fuses overlapping regions without introducing grid artifacts. Finally, a Patch-Wise Spectral Fusion module combines the outputs of both branches in the frequency domain, merging low-frequency structural information from the global branch with high-frequency detail components from the local branch, resulting in high-fidelity and high-resolution generation.}
	\label{fig:fig3}
\end{figure*}

\begin{figure*}[htbp]
	\centering
	\includegraphics[width=1\textwidth, keepaspectratio]{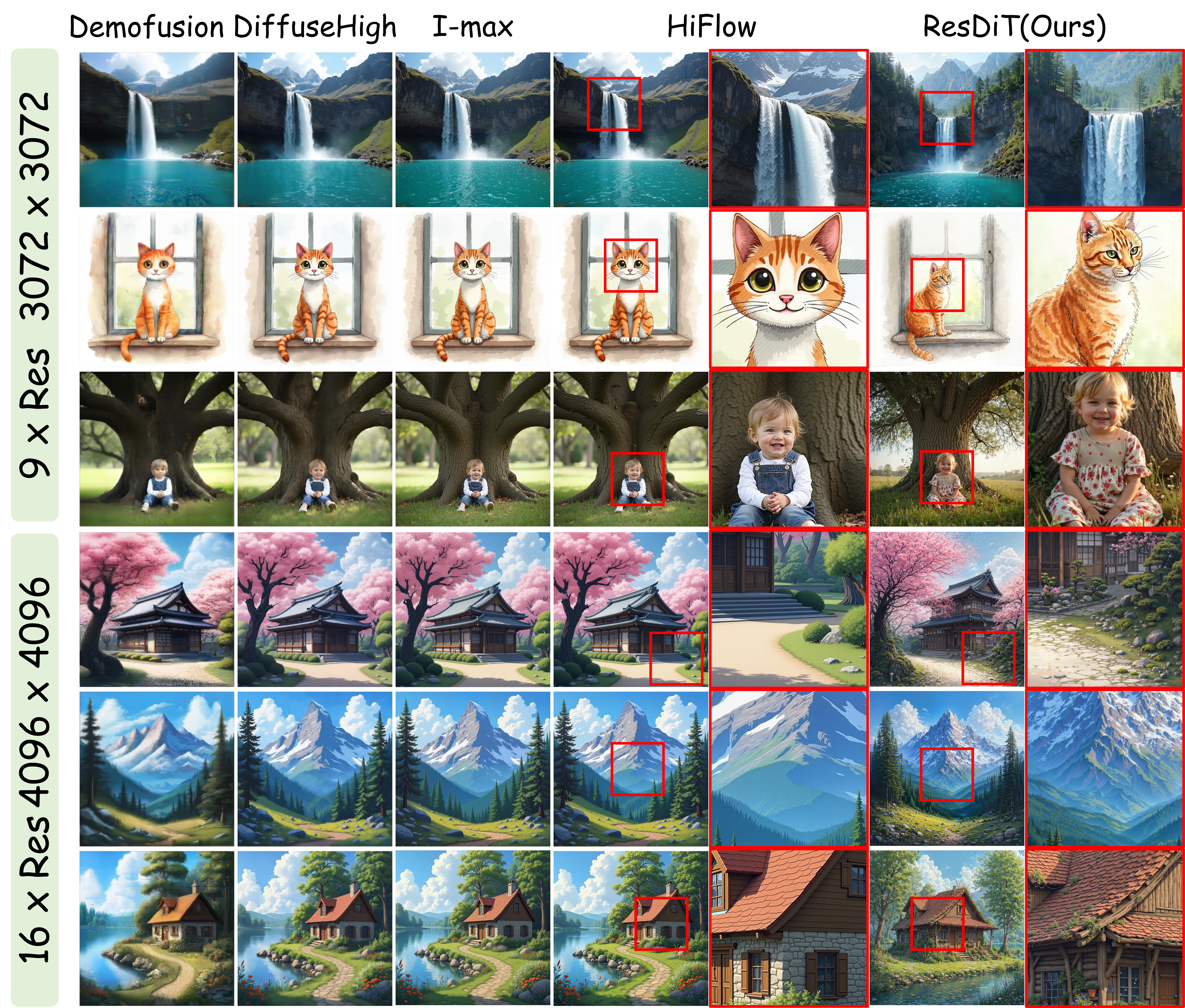}
	\caption{\textbf{Qualitative comparison with baselines.} ResDiT achieves a coherent global structure without relying on base resolution image information, while simultaneously delivering richer and more delicate local details in high-resolution outputs compared to existing methods. We further compare ResDiT with sota methods in terms of the capacity to generate fine-grained local details. \textbf{Best View ZOOM-IN.}}
	\label{fig:fig4}
\end{figure*}

\section{ResDiT}
Here, we propose ResDiT, a training-free method that scales resolutions of pre-trained Diffusion Transformers (DiTs) by exploiting their intrinsic generative properties. As illustrated in~\cref{fig:fig2}, ResDiT restructures the vanilla attention operation into two targeted branches. The first branch performs global attention with scaled position embedding to rectify the overall layout, while the second applies patch-level attention to recover fine-grained local details~\cref{sec:1}. By employing the proposed Minimum-Overlap Partitioning and Gaussian Weighting Splicing strategies~\cref{sec:2}, feature patches can be divided and seamlessly reunited without introducing grid artifacts.   Furthermore, ResDiT incorporates a Patch-Wise Spectral Fusion module to merge the two branches in the frequency domain, enabling a clean separation between layout-level and detail-level components and their effective integration~\cref{sec:3}.

\label{sec:method}

\subsection{Preliminaries}
\myparagraph{Positional Embedding (PE).} 
Positional embedding provides spatial priors for Transformer architectures by encoding coordinate information into feature representations. Among various designs, Rotary Position Embedding (RoPE)~\cite{su2024roformer} is a widely used scheme that encodes relative positions through rotation in the embedding space, and it has been adopted in recent T2I models such as FLUX. Since our method operates directly on positional indices, it is compatible with both RoPE and other positional encoding variants.

\myparagraph{Diffusion Transformer (DiT).}
DiT replaces the U-Net backbone with a pure Transformer architecture that processes image features as token sequences. 
Given image tokens $\mathbf{x} \in \mathbb{R}^{s \times c}$,  the core computation in each Transformer block is the self-attention operation, defined as:
\[
\mathbf{Q} = \mathbf{x} W_Q,\qquad
\mathbf{K} = \mathbf{x} W_K,\qquad
\mathbf{V} = \mathbf{x} W_V,
\]
\begin{equation}
	\mathrm{Attn}
	= \mathrm{softmax}\!\left(
	\frac{\mathbf{Q}\mathbf{K}^{\top}}{\sqrt{d_k}}
	\right)\mathbf{V}.
\end{equation}
This standard attention formulation serves as the basis for our two-branch attention restructuring in ResDiT.

\subsection{Position Embedding Rectification}\label{sec:1}
The original position embedding (PE) is not well-suited for high-resolution (HR) generation. As we discussed in~\cref{fig:fig2}, we present two specified PE rectification patterns for each attention branch.

\myparagraph{PE Scaling (Interpolation).} For the first branch, to retain a consistent global layout, we interpolate the position embedding from the base resolution to the target high resolution. This constrains the position information of the entire feature map to the range that the pre-trained model is familiar with, helping it preserve the global structural skeleton of the generated image. Formally, let $H$, $W$ be the height and width of a high-resolution feature map, the corresponding 2D position indices $p_{h} $ and $ p_{w}$ are defined as:
\begin{equation}
	(p_{h}, p_{w}) \in \{0, 1, \dots, H\!-\!1 \} \times \{0, 1, \dots, W\!-\!1\},
\end{equation}
We then scale the position indices to the model’s training range:
\begin{equation}
	(p_{h}, p_{w}) \in \{\tfrac{0}{s_{h}}, \tfrac{1}{s_{h}}, \dots, \tfrac{H\!-\!1}{s_{h}}\} \times \{\tfrac{0}{s_{w}}, \tfrac{1}{s_{w}}, \dots, \tfrac{W\!-\!1}{s_{w}}\},
\end{equation}
where $s_{h} = H/h$, $s_{w} = W/w$ and $h$, $w$ represent the height and width at the training resolution. The scaled indices are then used to compute positional embeddings.

\myparagraph{Patch-wise Independent PE.}  In the second branch, to align with patch-wise attention, each feature patch is assigned an independent positional embedding to maintain strong detail generation and reinforce local fidelity. 

\begin{table*}[t!]
	\centering
	\small
	\renewcommand{\arraystretch}{0.9} 
	\setlength{\tabcolsep}{5pt} 
	
	\begin{tabular}{ll cccccc} 
		\toprule
		\textbf{Resolution} & \textbf{Method} & \textbf{KID}$\downarrow$ & \textbf{KID$_p$}$\downarrow$ & \textbf{IS}$\uparrow$ & \textbf{IS$_p$}$\uparrow$ & \textbf{CLIP}$\uparrow$ & \textbf{User Study}$\uparrow$ \\
		\midrule
		
		\multirow{5}{*}{3072 $\times$ 3072} 
		& Demofusion~\cite{du2024demofusion} & 0.0211 & 0.0342 & 12.20 & 10.21 & 31.92 & 3.1 \\
		& DiffuseHigh~\cite{kim2025diffusehigh} & 0.0195 & 0.0213 & 12.61 & 10.13 & 32.74 & 4.2 \\
		& I-Max~\cite{du2024max} & 0.0192 & 0.0207 & \textbf{12.96} & 10.48 & 32.73 & 4.2 \\
		& HiFlow~\cite{bu2025hiflow} & \underline{0.0190} & \textbf{0.0194} & 12.87 & \underline{10.67} & \underline{32.76} & \underline{4.6} \\ 
		& ResDiT(Ours) & \textbf{0.0189} & \underline{0.0199} & \underline{12.91} & \textbf{10.87} & \textbf{32.85} & \textbf{4.8} \\ 
		
		\midrule 
		
		\multirow{5}{*}{4096 $\times$ 4096} 
		& Demofusion~\cite{du2024demofusion} & 0.0236 & 0.0374 & 10.56 & 9.23 & 30.41 & 2.1 \\
		& DiffuseHigh~\cite{kim2025diffusehigh} & 0.0215 & 0.0298 & 11.40 & 9.81 & 32.70 & 3.9 \\
		& I-Max~\cite{du2024max} & \underline{0.0208} & 0.0275 & \textbf{11.79} & 9.95 & 32.69 & \underline{4.0} \\
		& HiFlow~\cite{bu2025hiflow} & \textbf{0.0203} & \textbf{0.0245} & \underline{11.65} & \textbf{10.12} & \textbf{32.74} & \textbf{4.3} \\ 
		& ResDiT(Ours) & 0.0217 & \underline{0.0252} & 11.46 & \underline{9.97} & \underline{32.71} & \textbf{4.3} \\ 
		
		\bottomrule
	\end{tabular}%
	
	\caption{\textbf{Quantitative comparisons with baselines.} The best results are highlighted in \textbf{bold}, and the second-best results are \underline{underlined}. Compared to approaches that rely on base-resolution images for high-resolution generation, ResDiT achieves competitive, near state-of-the-art performance at 3072 × 3072 resolution. A slight performance drop is observed at 4096 × 4096, which is further analyzed in the experimental section.}
	\label{tab:tab1}
\end{table*}

\subsection{Patch Partitioning \& Splicing}\label{sec:2}
As discussed in~\cref{fig:fig2}, global attention over a high-resolution feature map often causes blurred textures and loss of fine details because the model is forced far beyond the spatial scale it was trained on. A natural solution is to confine attention to patch-level regions matching the training resolution. However, naïvely partitioning the feature map into a rigid grid introduces visible seams and grid artifacts along patch boundaries. To overcome this, we introduce partitioning and splicing strategies that enable patch-level attention while maintaining smooth, artifact-free continuity across patches.

\myparagraph{Minimum-Overlap Partitioning.}Non-overlapping feature patches can cause undesired and non-negligible image discontinuities, as features near patch boundaries lack contextual information from adjacent regions. To address this issue, we adopt a minimum-overlap partitioning strategy, where neighboring patches slightly overlap and thus share boundary context, smoothing feature transitions and reducing visible artifacts. 

Concretely, along a single spatial axis of length $H$ with patch size $h$, we choose an integer $N$ such that $N > H/h$, and place the $k$-th patch ($k = 1,\dots,N$) with starting index
\begin{equation}
	t_k = \frac{(k-1)(H - h)}{N - 1}.
\end{equation}
In this way, the first patch starts at $t_1 = 0$, the last patch ends at $t_N + h = H$, and the stride between neighboring patches is smaller than $h$, ensuring both full coverage of the axis and a positive overlap between adjacent patches with only a small number of partitions.

\myparagraph{Gaussian Weighting Splicing.}When splicing the overlapped regions of partitioned patches, instead of treating each patch equally, we apply a Gaussian weighting strategy to integrate the overlapping features in a weighted manner. It further alleviates boundary artifacts by enabling smoother feature transitions between neighboring patches. Formally, for a token located at $\bm{p}$ within an overlap region, let $\mathcal{W}(\bm{p})$ denote the set of attention windows that cover $\bm{p}$. For each patch $i \in \mathcal{W}(\bm{p})$, we assign a Gaussian weight:
\begin{equation}
	w_i(\bm{p})
	= \exp\!\Bigl(
	-\frac{\lVert \bm{p} - \bm{c}_i \rVert_2^{\,2}}{2\sigma^{2}}
	\Bigr),
\end{equation}
where $ \bm{c}_i $ is the centre of patch $i$ and $\mathcal{W}(\bm{p}) \in (0, 1]$. The final fused feature of token $\bm{p}$ is obtained:
\begin{equation}
	\bm{f}(\bm{p})
	= \frac{\sum_{i\in\mathcal{W}(\bm{p})} w_i(\bm{p})\,\bm{f}_i(\bm{p})}
	{\sum_{i\in\mathcal{W}(\bm{p})} w_i(\bm{p}) }.
\end{equation}

\subsection{Patch-wise Spectral Fusion}\label{sec:3}
To effectively integrate the outputs of the two branches, we first note that they contribute complementary information: the global branch captures reliable low-frequency layout structures, while the local branch excels at high-frequency details. This motivates us to perform fusion in the frequency domain, where such components can be cleanly separated and recombined. Specifically, we suppress high-frequency components in the global branch and low-frequency components in the local branch, enabling the two sources to merge in a naturally complementary manner.

Our fusion operates at the patch level, as the frequency composition varies significantly across spatial regions—high frequencies dominate textured or edge-rich areas, while low frequencies are more prevalent in smooth or homogeneous regions. Patch-wise spectral fusion allows the model to adaptively modulate frequency components for each region, yielding more flexible and effective integration than applying a global frequency filter.

Specifically, we partition the global output feature $\mathbf{x}_{g}$ from the first branch using the same minimum-overlap partitioning strategy:
\begin{equation}
	\mathbf{x}_{g} = \{\mathbf{x}_{g}^{1}, \dots, \mathbf{x}_{g}^{i}, \dots\},
\end{equation}
which correspond one-to-one with the local patch outputs. For each aligned patch pair $(\mathbf{x}_{g}^{i}, \mathbf{x}_{l}^{i})$, we apply the Fast Fourier Transform~\cite{brigham1988fast} (FFT) to obtain their spectral representations:
\begin{equation}
	\hat{\mathbf{x}}_{g}^{i} = \mathcal{F}(\mathbf{x}_{g}^{i}), \quad
	\hat{\mathbf{x}}_{l}^{i} = \mathcal{F}(\mathbf{x}_{l}^{i}).
\end{equation}
We reconstruct the spatial-domain patch by applying the inverse FFT to the fused spectrum:
\begin{equation}
	\mathbf{x}^{i} = \mathcal{F}^{-1}\big(\mathbf{M} \odot \hat{\mathbf{x}}_{g}^{i} + (1 - \mathbf{M}) \odot \hat{\mathbf{x}}_{l}^{i}\big),
\end{equation}
where $\mathbf{M}$ is a binary mask that filters the frequency spectrum, and $\mathcal{F}^{-1}$ denotes the inverse FFT that maps features back to the spatial domain.

%% file: sec/4_exp.tex
\section{Experiments}

\begin{figure*}[htbp]
	\centering
	\includegraphics[width=0.9\textwidth, keepaspectratio]{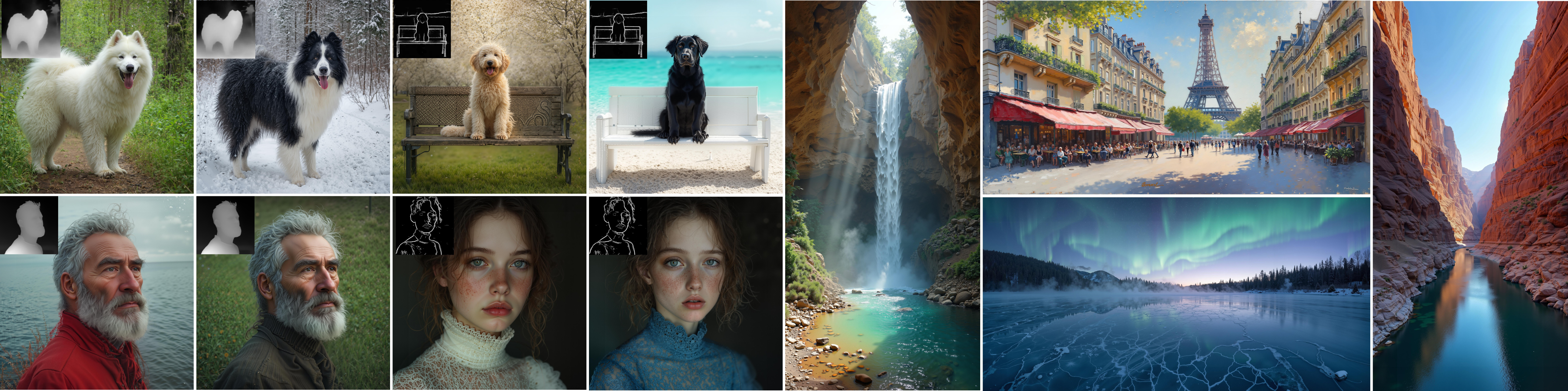}
	\caption{ResDiT seamlessly integrates with ControlNet, enabling precise structure-controlled generation of the images at resolutions of 3072 × 3072. Furthermore, ResDiT supports arbitrary aspect ratios, the images at resolutions of 2048 × 4096 and 4096 × 2048.}
	\label{fig:fig5}
\end{figure*}

\subsection{Implementation Details}

\myparagraph{Experimental settings.}If not specified, we deploy ResDiT on FLUX.1-dev~\cite{flux2024}, an advanced open-source model based on DiT architecture. The sampling steps are set to 35, and the guidance scale is set to 3.5. Following prior work~\cite{lu2024freelong,lu2025freelong++}, we set the normalized frequency cutoff of the spectral mask to 0.2. Recent research has observed the coarse-to-fine nature of diffusion denoising~\cite{yi2024towards}; inspired by this, we use a global branch for the first 10 timesteps, a local branch for the last 15 timesteps, and Patch-wise Spectral Fusion for the remaining steps. This strategy is similar to those proposed in prior work~\cite{he2023scalecrafter,zhou2025exploring}. All experiments are conducted on RTX 4090 GPU.

\myparagraph{Baselines.}We compare our method with training-free high-resolution generation approaches, Demofusion~\cite{du2024demofusion}, DiffuseHigh~\cite{kim2025diffusehigh}, I-Max~\cite{du2024max} and HiFlow~\cite{bu2025hiflow}. They all adopt a two-stage paradigm: first generate a base-resolution image and then perform high-resolution extrapolation. All methods are evaluated using their official implementations.

\myparagraph{Evaluation.}We collect 500 high-quality captions and generate images corresponding to each caption. We selected Kernel Inception Distance~\cite{binkowski2018demystifying} (KID), Inception Score~\cite{salimans2016improved} (IS), and CLIP Score~\cite{radford2021learning} as our evaluation metrics. KID measures the similarity between the generated high-resolution images and the original resolution images. IS assesses the diversity and definition of the generated images, and CLIP Score represents the prompt-following capability. The KID is calculated between generated images and 2K real high-quality images sourced from LAION-Aesthetics-v2 6.5plus~\cite{schuhmann2022laion}. To further provide the concrete evaluation, we also adopted patch KID and patch IS as our evaluation metrics. Besides, we conduct a User Study for further evaluation. Twenty participants independently rated images on a scale from 1 to 5 for their image visual quality based on 40 randomly selected prompts per method. The average user scores are reported.

\subsection{Comparison to State-of-the-Art Methods}
\myparagraph{Qualitative comparison.}We qualitatively compare our method with baseline approaches in \cref{fig:fig4}. They all rely on base-resolution images as guidance for high-resolution generation. Specifically, Demofusion preserves the overall image structure but introduces significant noise and severe detail loss. DiffuseHigh improves upon this by reducing artifacts, yet still struggles with fine-detail fidelity. I-max delivers richer details while maintaining structural accuracy, but suffers from localized blurring. The state-of-the-art HiFlow achieves superior consistency with base-resolution references, yielding visually sharper results, yet at the cost of over-smooth textures and diminished fine details. For instance, zoomed regions reveal unnatural smoothness on the child’s face, impoverished tree trunk textures in the background, and blurred distant mountain contours. In contrast, our method simultaneously preserves accurate global structure, recovers rich fine-scale details, and delivers superior visual realism without requiring any base-resolution input as a reference.

\begin{figure*}[htbp]
	\centering
	\includegraphics[width=0.9\textwidth, keepaspectratio]{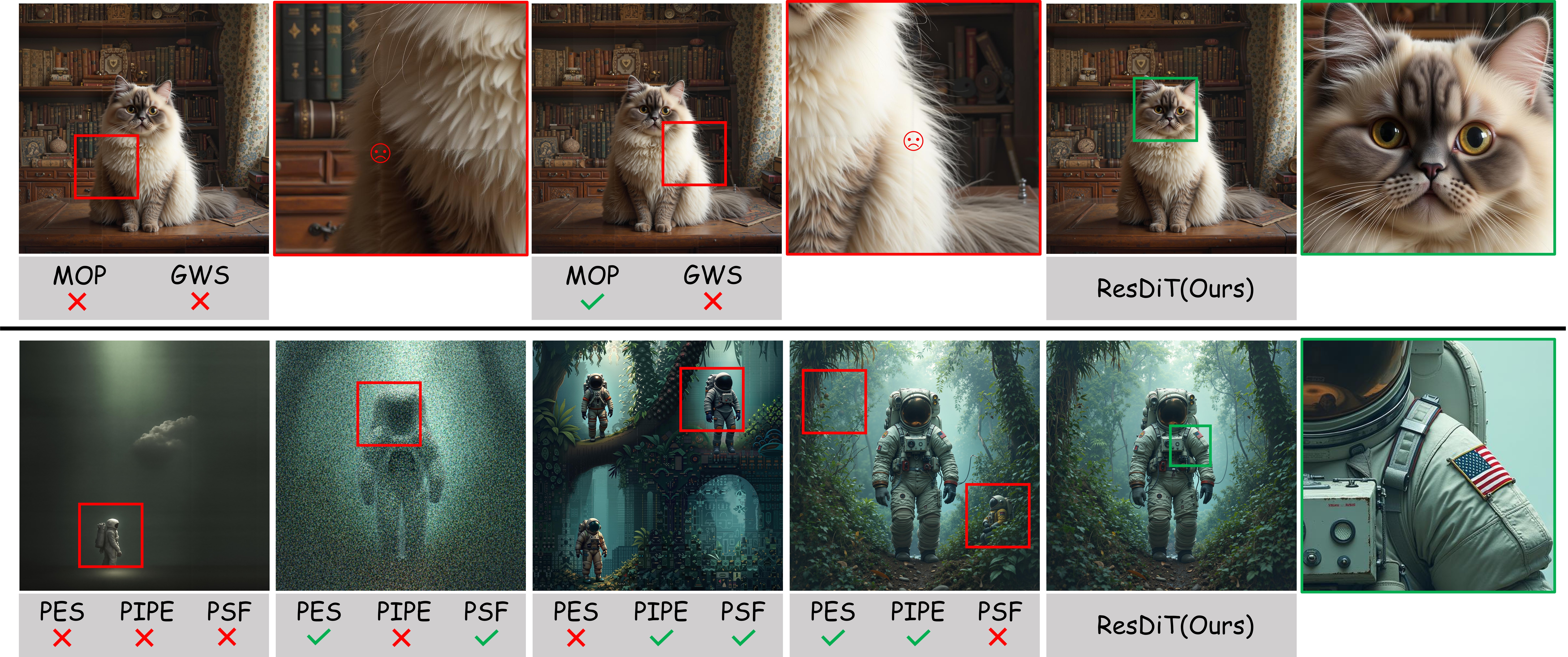}
	\caption{\textbf{Ablating each component of ResDiT.} Patch Partitioning \& Splicing confirms its role in preserving boundary fidelity and local detail. Patch Partitioning \& Splicing confirms its role in preserving boundary fidelity and local detail. Without PE Scaling (PES), the global structure becomes inconsistent. It ensures scalable coherence, while Patch-wise Independent PE (PIPE) adds rich fine details. Patch-wise Spectral Fusion (PSF) then synergistically combines both strengths, yielding superior generation quality.}
	\label{fig:fig6}
\end{figure*}

\myparagraph{Quantitative comparison.}We report the quantitative evaluation results in \cref{tab:tab1}. ResDiT achieves a notably high CLIP score, demonstrating strong image–text alignment. At a resolution of 3072 × 3072, it also attains superior KID and IS scores, reflecting improved image quality and diversity, benefiting from its ability to introduce rich fine-grained details while maintaining coherent global structure. However, at 4096 × 4096 resolution, a moderate drop in KID and IS scores is observed. We attribute this to the inherent difficulty of single-stage high-resolution generation. In contrast, most competing approaches adopt a two-stage strategy that heavily relies on base-resolution images generated by the model itself. As a result, their high-resolution outputs closely follow the distribution of the base-resolution images, leading to favorable scores. In comparison, ResDiT performs direct sampling in the high-resolution noise space without guidance from lower-resolution priors. While this design allows genuine high-resolution synthesis with richer and more diverse details, it also introduces a distributional shift from the original model’s generation space, which can lead to slightly inferior KID and IS metrics.

\myparagraph{Additional Qualitative Results.}\cref{fig:fig5} (left) illustrates ResDiT’s seamless compatibility with control modules such as ControlNet~\cite{zhang2023adding}, using depth and HED edge maps as structured priors under different textual prompts. The generated outputs show strong spatial alignment with the control inputs, confirming that ResDiT effectively preserves and enhances control consistency without compromising generation quality. The right side further demonstrates that ResDiT naturally supports arbitrary aspect ratios, while maintaining rational global layouts and delivering rich, high-fidelity local details, demonstrating its flexibility and robustness across diverse resolution configurations. More qualitative results are included in the appendix.

\subsection{Ablation Study}

\myparagraph{Position Embedding Rectification \& Patch-wise Spectral Fusion.}We validate the contribution of each component in our pipeline, as shown in the bottom part of \cref{fig:fig6}. We validated the respective contributions of PE Scaling (PES), Patch-wise Independent PE (PIPE), and Patch-wise Spectral Fusion (PSF). Removing PSF, replacing frequency-domain fusion with spatial-domain addition and averaging, led to repeated generation artifacts and increased image blurriness, confirming that frequency-domain fusion effectively leverages the complementary strengths of PES and PIPE while mitigating their drawbacks. Removing PES and reverting to the original positional encoding caused complete structural collapse, showing that PES is essential for modeling global structure at high resolution. When PIPE was removed, the global layout remained reasonable under PES, but fine details were severely degraded, demonstrating PIPE’s importance in enhancing local fidelity. Finally, removing all three resulted in both global structural collapse and local details distortion. Only the joint use of PES, PIPE, and PSF yields globally coherent and visually sharp results, highlighting their complementary roles in global layout control and fine-detail generation. Quantitative ablation experiments are included in the appendix.

\myparagraph{Patch Partitioning \& Splicing.}We conduct ablation study to evaluate the effectiveness of Minimum-Overlap Partitioning (MOP) and Gaussian Weighting Splicing (GWS). As shown in the upper part of \cref{fig:fig6}, removing both components, clear boundary artifacts emerge. Although our method maintains global information exchange, it inevitably introduces boundary discontinuities due to patch-level fusion and attention mechanisms. Using MOP alone greatly reduces segmentation artifacts, achieving results close to ours. However, for high-resolution generation, fine details remain crucial. Zooming into local regions reveals that artifacts persist, overlapping alleviates their visibility but increases their number and damages delicate details. In contrast, combining them effectively eliminates boundary artifacts, ensuring smooth transitions and consistent spatial coherence across global and local features.

\section{Conclusion}
We introduce ResDiT, a training-free framework that extends pre-trained Diffusion Transformers to high-resolution generation without relying on base-resolution images. We investigated intrinsic causes of high-resolution failures in DiTs and found that positional embeddings govern spatial layout while attention receptive-field scale controls detail fidelity. By restructuring the attention mechanism into global and local branches, ResDiT effectively preserves large-scale structural coherence while enriching fine-grained visual details. Extensive experiments demonstrate that ResDiT achieves competitive or superior performance at high resolutions, validating its effectiveness as a simple yet powerful solution for high-resolution diffusion generation.

%% file: sec/cvpr_sup.tex

\appendix

\twocolumn[
\begin{center}
	
	\Huge \textbf{Appendix}  
	\vspace{1cm}
	\small
	\renewcommand{\arraystretch}{0.9} 
	\setlength{\tabcolsep}{6pt} 
	
	\begin{tabular}{c ccc ccccc}
		\toprule
		\textbf{Resolution} & \textbf{PES} & \textbf{PIPE} & \textbf{PSF} &
		\textbf{KID}$\downarrow$ & \textbf{KID$_p$}$\downarrow$ &
		\textbf{IS}$\uparrow$ & \textbf{IS$_p$}$\uparrow$ &
		\textbf{CLIP}$\uparrow$ \\
		\midrule
		
		\multirow{5}{*}{3072×3072} 
		& $\times$ & $\times$ & $\times$ & 0.0836 & 0.1958 & 7.80 & 4.06 & 25.08 \\
		& $\times$ & \checkmark & \checkmark & 0.0906 & 0.0842 & 11.05 & 6.97 & 30.72 \\
		& \checkmark & $\times$ & \checkmark & 0.2218 & 0.4426 & 8.54 & 3.03 & 24.17 \\
		& \checkmark & \checkmark & $\times$ & 0.0227 & 0.0336 & 12.04 & 9.84 & 32.51 \\
		& \checkmark & \checkmark & \checkmark & \textbf{0.0189} & \textbf{0.0199} & \textbf{12.91} & \textbf{10.87} & \textbf{32.85} \\
		
		\midrule
		
		\multirow{5}{*}{4096×4096} 
		& $\times$ & $\times$ & $\times$ & 0.2703 & 0.3164 & 6.9 & 3.53 & 21.28 \\
		& $\times$ & \checkmark & \checkmark & 0.1334 & 0.1239 & 10.16 & 6.56 & 26.63 \\
		& \checkmark & $\times$ & \checkmark & 0.2208 & 0.3976 & 8.56 & 3.77 & 24.69 \\
		& \checkmark & \checkmark & $\times$ & 0.0488 & 0.0760 & 10.35 & 8.10 & 31.03 \\
		& \checkmark & \checkmark & \checkmark & \textbf{0.0217} & \textbf{0.0252} & \textbf{11.46} & \textbf{9.97} & \textbf{32.71} \\
		
		\bottomrule
	\end{tabular}
	
	\bigskip
	\captionof{table}{\textbf{Quantitative ablation study} of PE Scaling (PES) / Patch-wise Independent PE (PIPE) / Patch-wise Spectral Fusion (PSF) components at different resolutions. The best results are marked in \textbf{bold}.}
	\label{tab}
\end{center}
]

\section{Quantitative Ablation Study}

We conducted quantitative experiments to further validate the contributions of the three core components in our pipeline as shown in~\cref{tab}, using the same implementation details as described in the main text. As analyzed in the main text, omitting the PSF leads to partial blurring in the generated images and introduces repeated artifacts. The primary impact is on visual quality, which explains the slight performance drop compared with ResDiT in quantitative evaluations. In contrast, the other two ablations severely degrade image quality and yield significantly lower scores on quantitative metrics.

Notably, for some metrics, the performance without PIPE is even worse than that of direct generation. We believe this is because, although direct generation produces structurally disordered results, it still preserves some local semantic information. In comparison, removing PIPE yields roughly correct global structures but introduces substantial noise and artifacts across the entire image, which may heavily affect certain metrics.

\section{More Qualitative Results}
\begin{figure*}[p]
	\centering
	\includegraphics[width=0.85\textwidth, keepaspectratio]{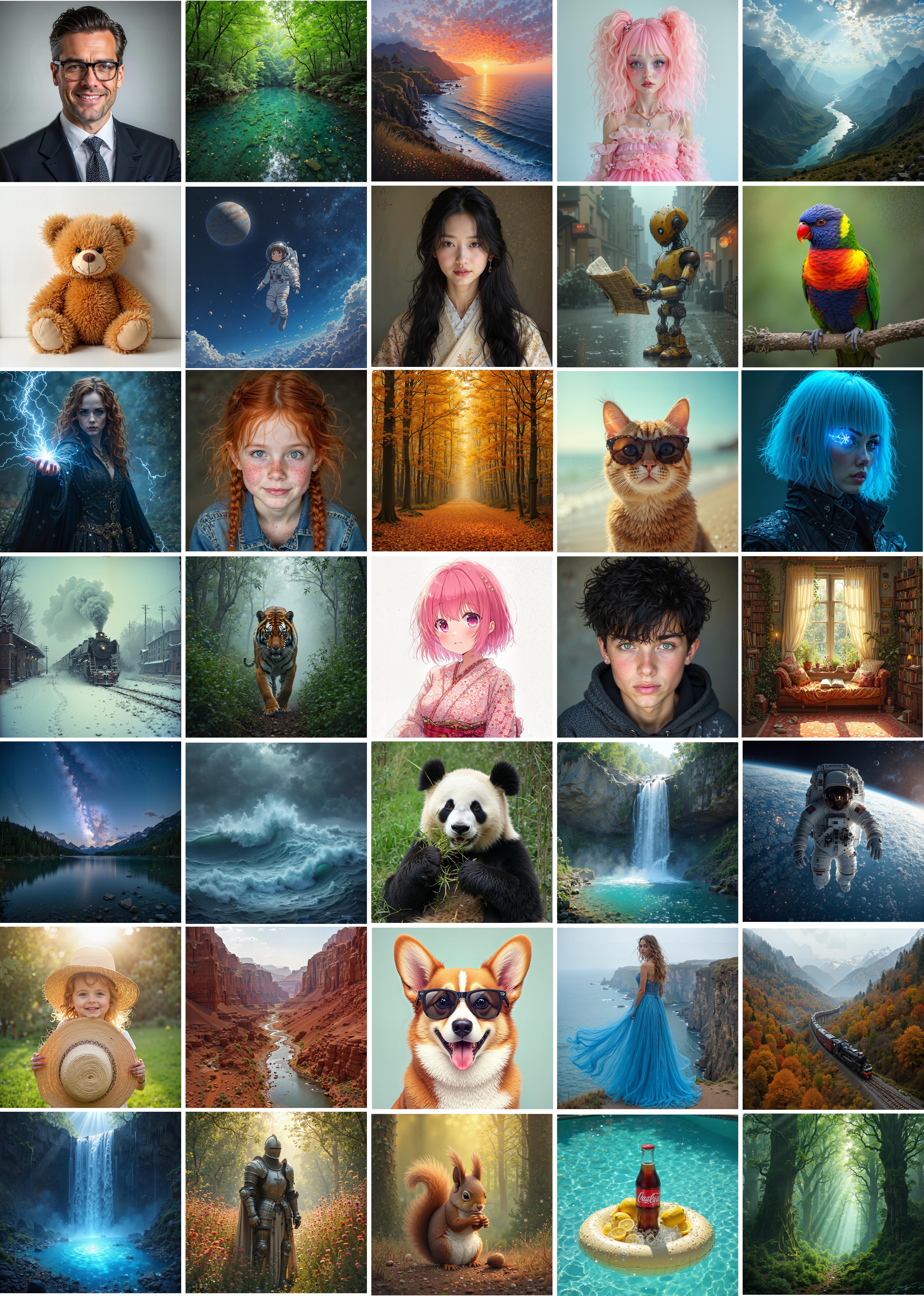}
	\caption{The 3072 × 3072 resolution image generated by ResDiT. \textbf{Best View ZOOM-IN.}}
	\label{3k}
\end{figure*}

\begin{figure*}[p]
	\centering
	\includegraphics[width=0.9\textwidth, keepaspectratio]{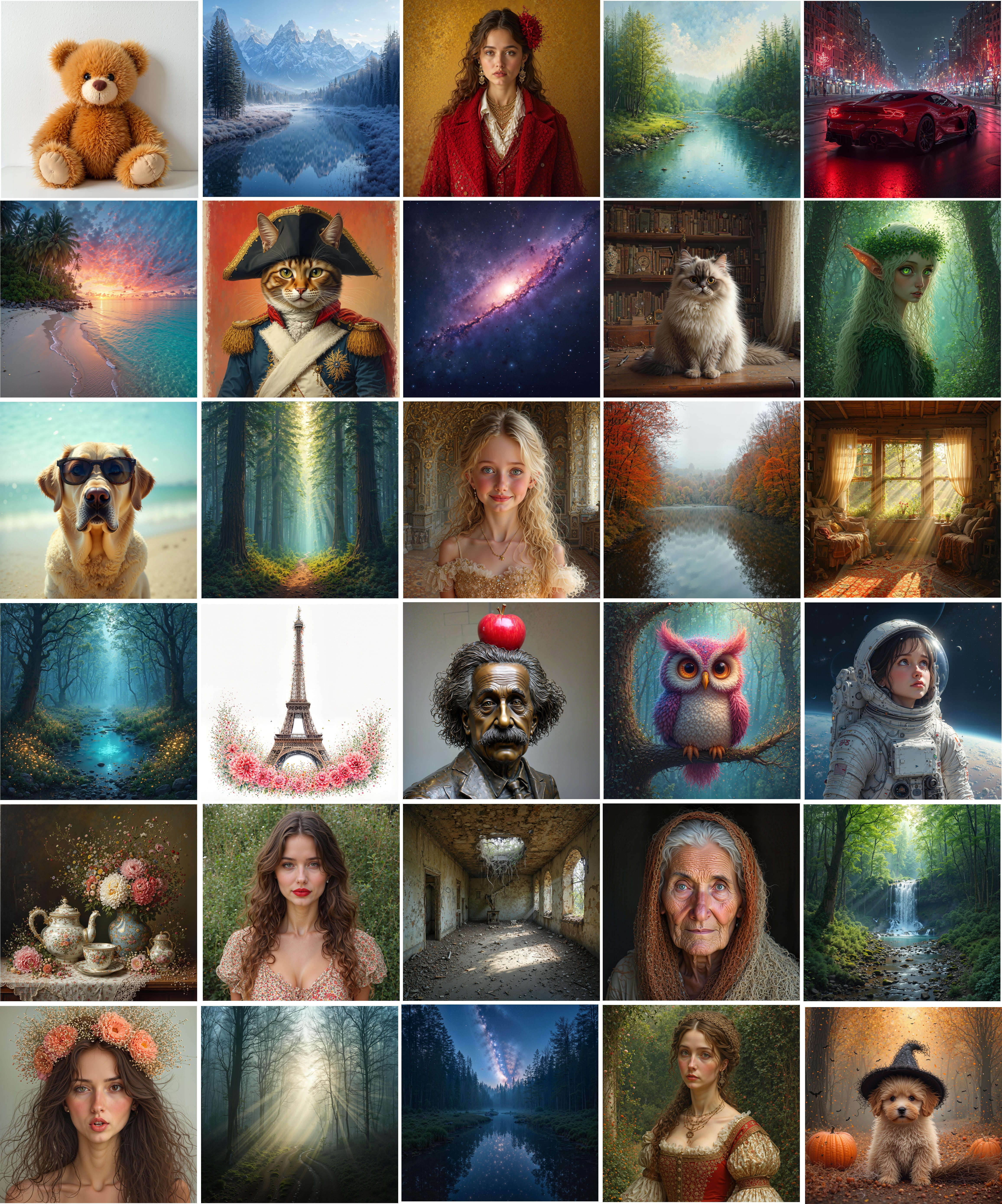}
	\caption{The 4096 × 4096 resolution image generated by ResDiT. \textbf{Best View ZOOM-IN.}}
	\label{4k}
\end{figure*}

We present additional ResDiT generation results below. Samples at resolution of 3072 × 3072 are shown in~\cref{3k}, and results at 4096 × 4096 resolution are displayed in~\cref{4k}.

